# Domain Adaptation through Synthesis for Unsupervised Person Re-identification

Sławomir Bąk*, Peter Carr*, Jean-François Lalonde†

Argo AI*, Université Laval†

**Abstract.** Drastic variations in illumination across surveillance cameras make the person re-identification problem extremely challenging. Current large scale re-identification datasets have a significant number of training subjects, but lack diversity in lighting conditions. As a result, a trained model requires fine-tuning to become effective under an unseen illumination condition. To alleviate this problem, we introduce a new synthetic dataset that contains hundreds of illumination conditions. Specifically, we use 100 virtual humans illuminated with multiple HDR environment maps which accurately model realistic indoor and outdoor lighting. To achieve better accuracy in unseen illumination conditions we propose a novel domain adaptation technique that takes advantage of our synthetic data and performs fine-tuning in a completely unsupervised way. Our approach yields significantly higher accuracy than semi-supervised and unsupervised state-of-the-art methods, and is very competitive with supervised techniques.

## 1 Introduction

Even over the course of just a few minutes, a person can look surprisingly different when observed by different cameras at different locations. Indeed, her visual appearance can vary drastically due to changes in her pose, to the different illumination conditions, and to the camera configurations and viewing angles. To further complicate things, she may be wearing the same shirt as another, unrelated person, and could thus easily be confused.

The task of person re-identification tackles the challenge of finding the same subject across a network of non-overlapping cameras. Most effective state-of-the-art algorithms employ supervised learning [1–6], and require thousands of labeled images for training. With novel deep architectures, we are witnessing an exponential growth of large scale re-identification datasets [1, 7]. Recent re-identification benchmarks have focused on capturing large numbers of identities, which allows the models to increase their discriminative capabilities [8].

Unfortunately, current re-identification datasets lack significant diversity in the number of lighting conditions, since they are usually limited to a relatively small number of cameras (the same person is registered under a handful of illumination conditions). Models trained on these datasets are thus biased to the illumination conditions seen during training. One can increase the model generalization by merging multiple re-identification datasets into a single dataset and training the network as joint single-task learning [8]. In this approach, the learned models show generalization properties but only upon fine-tuning [9]. This is because the merged datasets contain tens of different lighting conditions, which might not be sufficient to generalize. To apply the previously trained model to a new set of cameras, we need to annotate hundreds of subjects in each camera, which is a tedious process and does not scale to real-world scenarios.

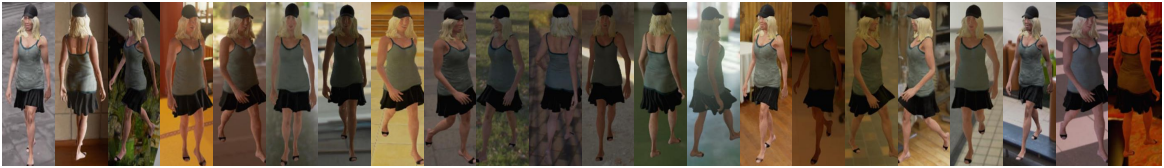

Fig. 1: Sample images from our **SyRI** dataset: the same 3D character rendered in various HDR environment maps. The dataset provides 100 virtual humans rendered in 140 realistic illumination conditions.

In this work, we introduce the ***Sy**nthetic Person **R**e-**I**dentification* (**SyRI**) dataset. Employing a game engine, we simulate the appearance of hundreds of subjects under different realistic illumination conditions, including indoor and outdoor lighting (see Fig. 1). We first carefully designed 100 virtual humans based on 3D scans of real people. These digital humans are then rendered using realistic backgrounds and lighting conditions captured in a variety of high dynamic range (HDR) environment maps. We use HDR maps as the virtual light source and background plate when rendering the 3D virtual scenes. With the increased diversity in lighting conditions, the learned re-identification models gain additional generalization properties, thus performing significantly better in unseen lighting conditions.

To further improve recognition performance, we propose a novel **three-step domain adaptation technique** that translates our dataset to the target conditions by employing cycle-consistent adversarial networks [10]. Since the cycle-consistent formulation often produces semantic shifts (the color of clothing may change drastically during translation), we propose an additional regularization term to limit the magnitude of the translation [11], as well as an additional masking technique to force the network to focus on the foreground object. The translated images are then used to fine-tune the model to the specific lighting conditions. In summary, our main contributions are:

– We introduce a new dataset with 100 virtual humans rendered with 140 HDR environment maps. We demonstrate how this dataset can increase generalization capabilities of trained models in unseen illumination conditions without fine-tuning.
– We improve re-identification accuracy in an unsupervised fashion using a novel three-step domain adaptation technique. We use cycle-consistency translation with a new regularization term for preserving identities. The translated synthetic images are used to fine-tune the re-identification model for a specific target domain.

## 2   Related work

**Person re-identification:** Most successful person re-identification approaches employ supervised learning [1,2,8,12–14]. This includes novel deep architectures and the debate as to whether the triplet or multi-classification loss is more effective for training re-identification networks [8,14,15]. Larger architectures have improved accuracy, but also increased the demand for larger re-identification datasets [7,16,17]. However all of these approaches require fine-tuning [9,16] to become effective in unseen target illumination conditions, which is infeasible for large camera networks. To overcome this scalability issue, semi-supervised and unsupervised methods have been proposed [18–20]. This includes transfer learning [19,21,22] and dictionary learning [23–25]. However, without labeled data, these techniques usually look for feature invariance, which reduces discriminativity, and makes the methods uncompetitive with supervised techniques.

**Synthetic data:** Recently, data synthesis and its application for training deep neural architectures has drawn increasing attention [11]. It can potentially generate unlimited labeled data. Many computer vision tasks have already been successfully tackled with synthetic data: human pose estimation [26], pedestrian detectors [27–29] and semantic segmentation [30,31]. The underlying challenge when training with synthetic visual data is to overcome the significant differences between synthetic and real image statistics. With increasing capacity of neural networks, there is a risk that the network will learn details only present in synthetic data and fail to generalize to real images. One solution is to focus on rendering techniques to make synthetic images appear more realistic. However, as the best renderers are not differentiable, the loss from the classifier cannot be directly back-propagated, thus leaving us with simple sampling strategies [28]. Instead, we take an approach closer to [11]: rather than optimizing renderer parameters, we cast the problem as a domain adaptation task. In our case, the domain adaptation performs two tasks simultaneously: (1) makes the synthetic images look more realistic and (2) minimizes the domain shift between the source and the target illumination conditions.

**Domain adaptation:** Typically, domain adaptation is a way of handling dataset bias [32]. Not surprisingly, domain adaptation is also used to minimize the visual gap betwen synthetic and real images [11]. Often this shift between distributions of the source and target domain is measured by the distance between the source and target subspace representations [33]. Thus, many techniques focus on learning feature space transformations to align the source and the target domains [21,34]. This enables knowledge transfer (*e.g.* how to perform a particular task) between the two domains. Recently, adversarial training has achieved impressive results not only in image generation [35], but also in unsupervised domain adaptation [36]. In this work, we are inspired by a recent approach for unsupervised image-to-image translation [10], where the main goal is to learn the mapping between images, rather than maximizing the performance of the model in particular task. Given our synthesized images and the domain translation, we are able to hallucinate labeled training data in the target domain that can be used for fine-tuning (adaptation).

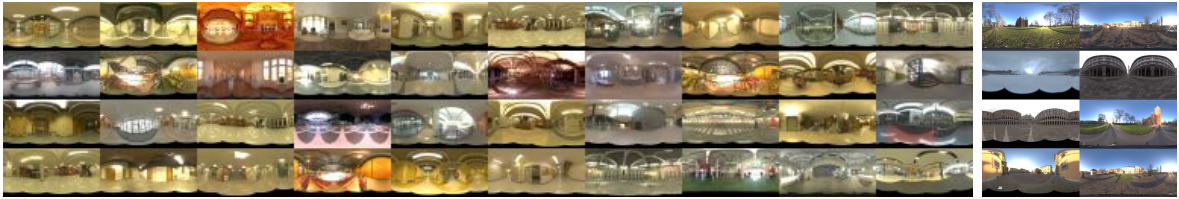

Fig. 2: **Example HDR environment maps used to relight virtual humans.** The environment maps capture a wide variety of realistic indoor (left) and outdoor (right) lighting conditions. The images have been tonemapped for display purposes with $\gamma = 2.2$. Please zoom-in for more details.
.

## 3   SyRI Dataset

Given sufficient real data covering all possible illumination variations, we should be able to learn re-identification models that have good generalization capabilities without the need for fine-tuning. Unfortunately, gathering and annotating such a dataset is prohibitive. Instead, we propose training with synthesized data. The underlying challenge is to create photo-realistic scenes with realistic lighting conditions. Rather than hand-crafting the illumination conditions, we use High Dynamic Range (HDR) environment maps [37]. These can be seen as 360° panoramas of the real world that contain accurate lighting information, and can be used to relight virtual objects and provide realistic backgrounds.

### 3.1   Environment maps

To accurately model realistic lighting, a database of 140 HDR environment maps was acquired. First, 40 of those environment maps were gathered from several sources online [1]. Further, we also captured an additional 100 environment maps. To do so, a Canon 5D Mark III camera with a Sigma 8mm fisheye lens was mounted on a tripod equipped with panoramic tripod head. 7 bracketed exposures were shot at 60° increments, for a total of 42 RAW photos per panorama. The resulting set of photos were automatically merged and stitched into a 22 f-stop HDR 360° environment map using the PTGui Pro commercial software. Our dataset represents a wide variety of indoor and outdoor environments, such as offices, theaters, shopping malls, museums, classrooms, hallways, corridors, etc. Fig. 2 shows example environment maps from our dataset.

### 3.2   3D virtual humans and animations

Our 3D virtual humans are carefully designed with **Adobe Fuse CC**[2] that provides 3D content, including body scans of real people with customizable body parts and clothing. We generate 100 character prototypes, where we customize body shapes, clothing, material textures and colors (see Fig. 3). These characters are then animated using rigs to obtain realistic looking walking poses.

---

[1] The following online ources were used: http://gl.ict.usc.edu/Data/HighResProbes/, http://dativ.at/lightprobes, http://www.unparent.com/photos_probes.html, http://www.hdrlabs.com/sibl/archive.html

[2] http://www.adobe.com/products/fuse.html

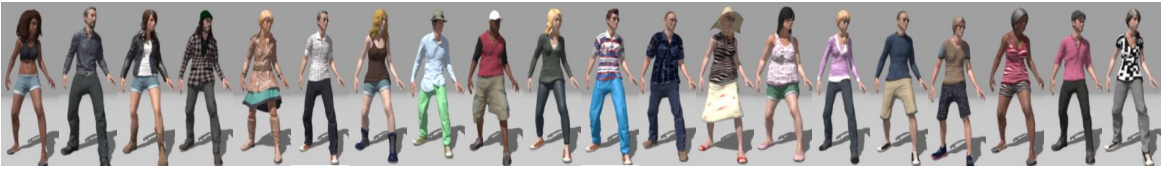

Fig. 3: Sample 3D virtual humans from **SyRI** dataset.

### 3.3 Rendering

We use **Unreal Engine 4**[3] to achieve real-time rendering speeds. To relight our 3D virtual humans, the HDR environment map is texture mapped on a large sphere surrounding the scene. This sphere is then used as a the sole light source (light emitter) to render the scene. We position a 3D character at the center of the sphere. The character is animated using either a male or female walking rig, depending on the model gender. We also add a rotation animation to acquire multiple viewpoints of each subject. The camera position is matched with existing re-identification datasets. Each subject is rendered twice under the same HDR map rotating the sphere about its vertical axis by two random angles. This effectively provides two different backgrounds and lighting conditions for each environment map. We render 2-second videos at 30 fps as the character is being rotated. In the end, we render $100\,(\text{subjects}) \times 140\,(\text{environment maps}) \times 2\,(\text{rotations}) \times 2\,(\text{seconds}) \times 30\,(\text{fps}) = 1,680,000$ frames. Both the rendered dataset as well as the Unreal Engine project that will allow a user to render more data are going to be made publicly available.

## 4 Method

We cast person re-identification as a domain adaptation problem, where the domain is assumed to be an illumination condition (*i.e.*, a camera-specific lighting). Our objective is to find an effective and unsupervised strategy for performing person re-identification under the target illumination condition.

For training, we assume we have access to $M$ real source domains $\mathbf{R} = \{R_1 \ldots R_M\}$, where each $R_m = \{x_i, y_i\}_{i=1}^{Z_{R_m}}$ consists of $Z_{R_m}$ real images $x_i$ and their labels $y_i$ (person's identity); and $N$ source synthetic domains $\mathbf{S} = \{S_1 \ldots S_N\}$, where each $S_n = \{s_i, y_i\}_{i=1}^{Z_{S_n}}$ consists of $Z_{S_n}$ synthetic images $s_i$ and their labels $y_i$ (3D character's identity). In our case $N \gg M$ as we have access to hundreds of different illumination conditions (see Sec. 3.1). Our ultimate goal is to perform re-identification in an unknown target domain $R_{M+1} = \{x_i\}_{i=1}^{Z_{R_{M+1}}}$ for which we do not have labels.

---
[3] https://www.unrealengine.com/

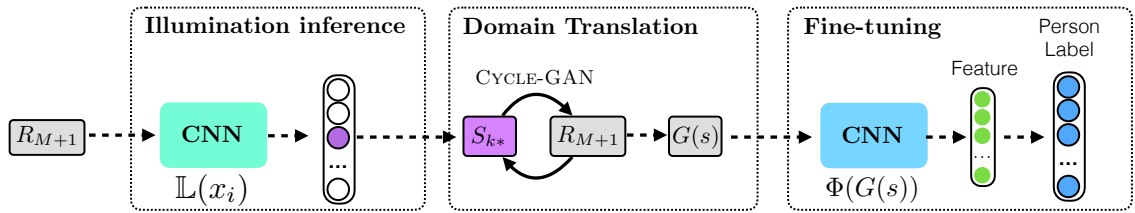

Fig. 4: **Unsupervised Domain Adaptation.** Given unlabelled input images from target domain $R_{M+1}$, we first select the closest synthetic domain $S_{k*}$ through illumination inference. Afterwards, images from the selected domain $S_{k*}$ are translated by $G : S_{k*} \to R_{M+1}$ to better resemble the input images in $R_{M+1}$. Finally, the translated synthetic images $G(s)$ along with their known identities are used to fine-tune the re-identification network $\Phi(\cdot)$.

### 4.1 Joint learning of re-identification network

We first learn a generic image feature representation for person re-identification. The feature extractor $\Phi(\cdot)$ is a Convolutional Neural Network (CNN) trained to perform multi-classification task, *i.e.* given a cropped image of a person, the CNN has to predict the person's identity. We propose to merge all domains **R** and **S** into a single large dataset and train the network jointly from scratch. We adopt the CNN model from [8]. To learn discriminative and generalizable features, the number of classes during training has to be significantly larger than the dimensionality of the last hidden layer (feature layer). In our case the training set consists of 3K+ classes (identities) and the feature layer has been fixed to 256 dimensions.

One could assume that with our new dataset, the pre-trained model should generalize well in novel target conditions. Although synthetic data helps (see Sec. 5.1), there is still a significant performance gap between the pre-trained model and its fine-tuned version on the target domain. We believe there are two reasons for this gap: (1) our dataset does not cover all possible illumination conditions, and (2) there is a gap between synthetic and real image distributions [11]. This motivates the investigation of domain adaptation techniques that can potentially address both issues: making the synthetic images looking more realistic, as well as minimizing the shifts between source and target illumination conditions.

### 4.2 Domain adaptation

We formulate domain adaptation as the following three-step process, as illustrated in Fig. 4.

1. **Illumination inference**: find the closest illumination condition (domain $S_{k*} \in \mathbf{S}$) for a given input $R_{M+1}$ (see Sec. 4.2).
2. **Domain translation**: translate domain $S_{k*}$ to $R_{M+1}$, by learning $G$, $G : S_{k*} \to R_{M+1}$ while preserving a 3D character's identity from $s \in S_{k*}$ (see Sec. 4.2).
3. **Fine-tuning**: update $\Phi(\cdot)$ with the translated domain $G(s)$ (see Sec. 4.2).

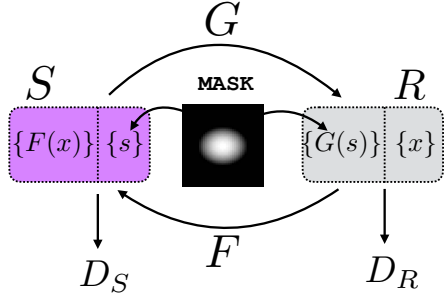

Fig. 5: **Semantic Shift Regularization**. The Cycle-GAN loss only applies to $F(G(s))$ and $G(F(x))$. There is no constraint on what $G()$ and $F()$ can do individually, which can result in drastic color changes. We incorporate an additional regularization loss requiring $s$ and $G(s)$ to be similar. The loss should only apply to the foreground (to preserve identity), since the target camera may have a very different background than the synthetic data.

**Illumination inference** Domain adaptation is commonly called a visual dataset bias problem. Dataset bias was compellingly demonstrated in computer vision by the *name the dataset* game of Torralba and Efros [32]. They trained a classifier to predict which dataset an image originated from, illustrating that visual datasets are biased samples of the visual world. In this work, we employ a similar idea to identify the synthetic domain $S_{k*} \in \mathbf{S}$ that is closest to the target domain $R_{M+1}$. To do so, we train a CNN classifier that takes an input image and predicts which illumination condition the image was rendered with. In our case, the classifier has to classify the image into one of $N = 140$ classes (the number of different environment maps in our synthetic dataset). We used Resnet-18 [38] pretrained on ImageNet and fine-tuned to perform illumination classification. Given the trained classifier, we take a set of test images from $R_{M+1}$ and predict the closest lighting condition by

$$k^* = \underset{k \in \{1...N\}}{\arg\max} \sum_{i=1}^{Z_{R_{M+1}}} \Delta\big(\mathbb{L}(x_i), k\big),$$

$$\text{where } \Delta\big(\mathbb{L}(x_i), k\big) = \begin{cases} 1, & L(x_i) = k \\ 0, & \text{otherwise} \end{cases}. \quad (1)$$

Here, $k$ corresponds to domain class, $\mathbb{L}(x_i)$ is the class predicted by the CNN classifier and $\Delta$ is a counting function. We use this formulation to find $S_{k*}$: the synthetic illumination condition that is most similar to the target domain $R_{M+1}$ (*i.e.* requiring the minimum amout of domain shift). $S_{k*}$ will be used to translate images from $S_{k*}$ to $R_{M+1}$ while preserving each 3D character's identity.

**Domain translation** Given two domains $S$ and $R$ (for convenience we skip sub-indices here) and the training samples $s_i \in S$ and $x_i \in R$, our objective is to learn a mapping function $G : S \to R$. As we do not have corresponding pairs between our synthetic and real domains, $G$ is fairly unconstrained and standard procedures will lead to the well-known problem of mode collapse (all input images map to the same output image). To circumvent this problem, we adapt the technique of [10], where rather than learning a single mapping $G : S \to R$, we exploit the property that translation should be *cycle-consistent*. In other words there should exist the opposite mapping $F : R \to S$, where $G$ and $F$ are inverses of each other.

We train both mappings $G$ and $F$ simultaneously, and use two *cycle consistency losses* to regularize the training: $s \to G(s) \to F(G(s)) \approx s$ and $x \to F(x) \to G(F(x)) \approx x$. $G$ and $F$ are generator functions, where $G$ tries to generate images $G(s)$ that look similar to images from domain $R$, and $F$ generates images $F(x)$ that should look like images from domain $S$. Additionally, two adversarial discriminators $D_S$ and $D_R$ are trained, where $D_S$ tries to discriminate between images $\{s\}$ and translated images $\{F(x)\}$; and analogously $D_R$ aims to distinguish between $\{x\}$ and $\{G(s)\}$ (see Fig. 5).

The training objective contains *adversarial losses* [35] for both $G$ and $F$, as well as two *cycle consistency losses*. The *adversarial loss* for $G$ is defined as

$$\mathcal{L}_{GAN}(G, D_R, S, R) = \mathbb{E}_{x \sim p_{data}(x)}[\log D_R(x)] \\ + \mathbb{E}_{s \sim p_{data}(s)}[\log(1 - D_R(G(s)))], \quad (2)$$

and we can analogously define *adversarial loss* for $F$, *i.e.* $\mathcal{L}_{GAN}(F, D_S, R, S)$. Both *cycle consistency losses* can be expressed as

$$\mathcal{L}_{cyc}(G, F) = \mathbb{E}_{s \sim p_{data}(s)}[||F(G(s)) - s||_1] + \mathbb{E}_{x \sim p_{data}(x)}[||G(F(x)) - x||_1]. \quad (3)$$

The final objective is

$$\mathcal{L}_{CycleGAN}(G, F, D_S, D_R) = \mathcal{L}_{GAN}(G, D_R, S, R) + \mathcal{L}_{GAN}(F, D_S, R, S) \\ + \lambda_1 \mathcal{L}_{cyc}(G, F), \quad (4)$$

where $\lambda_1$ controls the relative importance of the *cycle consistency losses*.

**Semantic Shift Regularization** In the above formulation, there is no constraint that the color distribution of the generated image $G(s)$ should be close to instance $s$. With large capacity models, the approach can map the colors within $s$ to any distribution, as long as this distribution is indistinguishable from the emperical distribution within $R$ ($F(x)$ will learn the inverse mapping). In our application, the color of a person's shirt (*e.g.* red) can drastically switch under $G(s)$ (*e.g.* to blue) as long as $F(G(S))$ is able to reverse this process (see Fig. 8). This semantic shift corrupts the training data, since a synthetic image and its corresponding domain translated variant could look very different (*e.g.* the labels are not consistent). Semantic shift can occur because the *cycle-consistency* loss does not regulate the amount by which the domains can be shifted.

As mentioned in [10], one can adopt the technique from [39] and introduce an additional loss that forces the network to learn an identity mapping when samples from the target domain are provided as input to the generator, *i.e.*

$\mathcal{L}_{id}(G, F) = \mathbb{E}_{x \sim p_{data}(x)}[||G(x) - x||_1] + \mathbb{E}_{s \sim p_{data}(s)}[||F(s) - s||_1]$. Although, this loss helps to some degree, many subjects still exhibited drastic shifts in appearance.

Alternatively, we can integrate the loss from [11] which ensures the translated synthetic image is not too different from the original synthetic image *i.e.* $\mathcal{L}_{Ref}(G) = \mathbb{E}_{s \sim p_{data}(s)}[||G(s) - s||_1]$. We found this loss often leads to artifacts in the translated synthetic images, since the regularization does not distinguish between background/foreground. In practice, only the appearance of the person needs to be preserved. The background of synthetic image could be very different than what appears in the real images captured by the target camera.

To circumvent this issue, we apply a masking function which forces the network to focus on the foreground region

$$\mathcal{L}_{Mask}(G) = \mathbb{E}_{s \sim p_{data}(s)}\left[ \left\| (G(s) - s) * \mathbf{m} \right\|_1 \right], \quad (5)$$

where $\mathbf{m}$ is a mask that encourages the mapping to preserve the appearance only near to the center (see Fig. 5). Because re-identification datasets have well cropped images, the foreground region is typically in the middle of the bounding box, with the background around the periphery. Therefore, we pre-define a soft matte that resembles a 2D Gaussian kernel.

Our full objective loss is

$$\begin{aligned}\mathcal{L}_{our}(G, F, D_S, D_R) = &\mathcal{L}_{GAN}(G, D_R, S, R) + \mathcal{L}_{GAN}(F, D_S, R, S) \\ &+ \lambda_1 \mathcal{L}_{cyc}(G, F) + \lambda_2 \mathcal{L}_{id}(G, F) \\ &+ \lambda_3 \mathcal{L}_{Mask}(G),\end{aligned} \quad (6)$$

where $\lambda_1 = \lambda_2 = 10$ and $\lambda_3 = 5$ in our experiments. Figure 6 illustrates sample results of the domain translation process.

**Fine-Tuning** Given our re-identification network (see Sec. 4.1), we can fine-tune its feature extraction process to specialize for images generated from $G(s)$, which is our approximation of data coming from target domain (test camera). In practice, when we need to fine-tune our representation to a set of cameras, for every camera we identify its closest synthetic domain $S_{k*}$ through our illumination inference, and then use it to learn a generator network that can transfer synthetic images to the given camera domain. The transferred synthetic images $G(s) : s \in S_{k*}$ are then used for fine-tuning the re-identification network, thus maximizing the performance of $\Phi(G(s))$.

## 5   Experiments

We carried out experiments on 5 datasets: **VIPeR** [40], **iLIDS** [41], **CUHK01** [42], **PRID2011** [43] and **Market-1501** [7]. To learn a generic feature extractor we used two large scale re-identification datasets: **CUHK03** [1] and

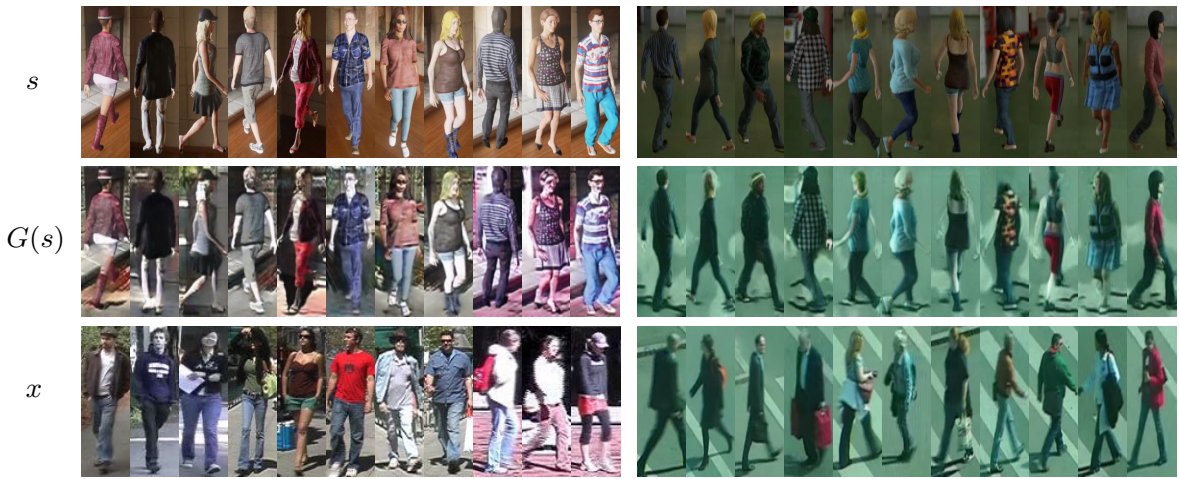

Fig. 6: **Domain translation results** for VIPeR (left) and PRID (right) datasets. From top to bottom: domain images $s \in S_{k*}$, translated images $G(s)$, target images $x \in R_{M+1}$.

**DukeMTMC4ReID** [17,44], and our **SyRI** dataset. Re-identification performance is reported using rank-1 accuracy of the CMC curve [40], which is the probability of finding the correct match in the first rank.

**Datasets:** VIPeR contains 632 image pairs of cropped pedestrians captured by two outdoor cameras. Large variations in lighting conditions, in background and in viewpoint are present. PRID2011 consists of person images recorded from two non-overlapping static surveillance cameras. Characteristic challenges of this dataset are extreme illumination conditions. There are two camera views containing 385 and 749 identities, respectively. Only 200 people appear in both cameras. i-LIDS consists of 476 images with 119 individuals. The images come from airport surveillance cameras. This dataset is challenging due to many occlusions, luggage and crowds. CUHK01 consists of 3,884 images of 971 identities. There are two images per identity, per camera. The first camera captures the side view of pedestrians and the second camera captures the front or back view. Market-1501 contains 1501 identities, registered by at most 6 cameras. All the images were cropped by an automatic pedestrian detector, resulting in many inaccurate detections.

**Evaluation protocol:** We generate probe/gallery images accordingly to the settings in [8]: VIPeR: 316/316; CUHK01: 486/486; i-LIDS: 60/60; and PRID2011: 100/649, where we follow a single shot setting [45]. For Market-1501 we employ the protocol from [46], where 750 identities are used for testing in a single query setting.

### 5.1 Generalization properties

In this experiment, we train two feature extractors: one with only real images **R** containing CUHK03 and DukeMTMC4ReID images (in total 3279 identities); and the other one with both real and our synthetic images **R** + **S** (our

| Method | VIPeR | CUHK01 | iLIDS | PRID | Market |
|---|---|---|---|---|---|
| State-of-the-Art | 38.5 [47] | **57.3** [46] | 49.3 [48] | 34.8 [47] | 58.2 [47] |
| **R** | 32.3 | 41.6 | 51.0 | 7.0 | 44.7 |
| **R + S** | 36.4 | 49.5 | 54.8 | 15.0 | 54.3 |
| **R + S*** | 49.4 | 71.4 | 63.2 | 65.0 | 83.9 |
| CycleGan | 37.0 | 49.9 | 53.9 | 33.0 | 55.4 |
| CycleGan+$\mathcal{L}_{id}$ | 39.9 | 54.0 | 55.9 | 40.0 | 63.1 |
| CycleGan+$\mathcal{L}_{Ref}$ | 41.1 | 48.4 | 56.1 | 28.0 | 57.5 |
| Ours | **43.0** | 54.9 | **56.5** | **43.0** | **65.7** |

Table 1: **CMC rank-1 accuracy**. The base model **R** is only trained on real images from auxiliary re-identification datasets. Adding synthetic images **S** improves the performance. Fine-tuning (indicated with *) to the training data of a specific dataset implies the maximum performance that could be expected with the correct synthetic data. Adapting the synthetic data to the target domain leads to significant gains, depending on the combination of semantic shift regularizations. Compared with state-of-the-art unsuperivsed techniques, our approach yields significantly higher accuracy on 4 of the 5 datasets. We achieve competitive performance to state-of-the-art on CUHK01.

**SyRI** dataset provides additional 100 identities but under 140 illumination conditions, for a total of 3379 identities). For **S** we used 4 randomly sampled images per illumination condition per identity, which results in 56,000 images ($4 \times 140 \times 100$). Table 1 reports the performance comparison of these models on various target datasets. First, we evaluate the performance of the models directly on the target datasets without fine-tuning (fully unsupervised scenario, compare rows **R** and **R + S**, respectively) . Adding our synthetic dataset significantly increases the re-identification performance. The row marked with * are the results after fine-tuning on the actual target datasets (*e.g.* in VIPeR column we fine-tune the model only on VIPeR dataset). It represents the maximum performance we expect to achieve if we could somehow hallucinate the perfect set of domain translated synthetic training images. These results indicate that the performance of supervised methods (using additional real data directly from the target domain) is still significantly better than unsupervised methods using domain adaptation. Interestingly, although adding our synthetic dataset doubled the performance on PRID2011, the lighting conditions in this dataset are so extreme that the gap to the supervised model is still significant. Similar findings have been reported in [9, 47].

## 5.2 Illumination inference

We carry out experiments to evaluate the importance of selecting proper illumination conditions. To do so, we compare the proposed *illumination inference* (sec. 4.2) to a random selection of the target illumination condition $S_{k*}$. After the illumination condition is selected, the proposed domain translation is applied. Table 2 illustrates the comparison on multiple dataset. We report minimum performance obtained by random procedure (MIN), the average across 10 experiments (Random), and the average using our illumination inference. The results demonstrate that reasoning about illumination greatly facilitates domain translation, and thereby helps in improving the recognition performance.

| Method | VIPeR | CUHK01 | iLIDS | PRID | Market |
|---|---|---|---|---|---|
| **R + S** | 36.4 | 49.5 | 54.8 | 15.0 | 54.3 |
| Min | 35.2 | 50.4 | 55.1 | 29.0 | 58.1 |
| Random | 38.9 | 51.2 | 56.1 | 36.0 | 60.9 |
| **Our** | **43.0** | **54.9** | **56.5** | **43.0** | **65.7** |

Table 2: Impact of illumination inference. The selection of the right illumination condition for the domain translation improves the recognition performance.

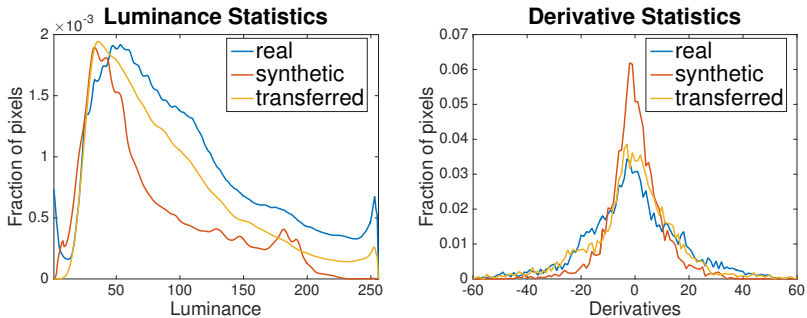

Fig. 7: **Comparison of image statistics.** Domain translation decreases the gap between synthetic and real image statistics.

### 5.3 Image statistics of SyRI

The effect of domain translation is reflected in the underlying image statistics (see Fig. 7). The statistics of real and synthetic images are derived from a single camera from the VIPeR dataset and its corresponding camera in our SyRI dataset (selected by illumination inference). After passing through the generator function learned during domain translation ($G(s)$), the statistics of the translated images are much closer to the statistics of real images.

### 5.4 Domain adaptation

Table 1 reports the performance of *CycleGAN* with different regularization terms. Domain translation without any regularization term between $s$ and $G(s)$ can deteriorate performance (compare **R + S** and CycleGAN for iLIDS). We suspect this is due to the previously mentioned semantic shift (see Fig. 8). Adding identity mapping $\mathcal{L}_{id}$ makes significant improvement on both visual examples and re-identification performance. Replacing $\mathcal{L}_{id}$ with $\mathcal{L}_{Ref}$ can lower performance and tends to produce artifacts (notice artificial green regions in Figure 8 for CUHK01). For CUHK01 and PRID datasets there are significant drops in the performance when using $\mathcal{L}_{Ref}$ regularization. Unlike [11], our images have distinct foreground/background regions. Background is not useful for re-identification, and it's influecen in the loss function should be minimial. Incorporating our mask makes significant improvements—especially for datasets where images are less tightly cropped, such as PRID. In this case, adding synthetic data improved performance from 7% to 15%. Our domain adaptation technique boosts the performance to 43.0% rank1-accuracy. We surpass the current state-of-the-art results by 8.2%.

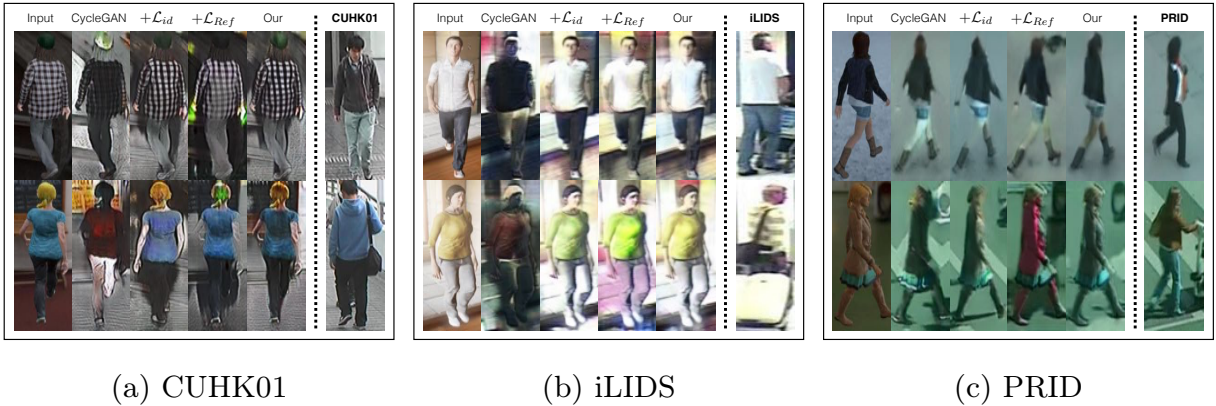

(a) CUHK01           (b) iLIDS           (c) PRID

Fig. 8: **Comparison of different regularization terms for translating synthetic images to a target domain.** Representative image pairs for CUHK01, iLIDS and PRID datasets have been selected. Notice that CycleGAN without any regularization tends to have semantic shifts, *e.g.* for CUHK01 blue color of the t-shirt changed to red.

### 5.5  Comparison with state-of-the-art methods

We divide the state-of-the-art approaches into unsupervised and supervised techniques as well as methods that employ hand-crafted features (including graph-learning GL [18] and transfer learning TL [18]) and embeddings learned with Convolutional Neural Newtworks (CNN) (including source identity knowledge transfer learning CAMEL [46] and attribute knowledge transfer TJ-AIDL [47]). Table 3 illustrates that: (1) our three-step domain adaptation technique outperforms the state-of-the-art unsupervised techniques—on 4 of the 5 datasets, we outperform the state-of-the-art results by large margins: 5.1%, 7.2%, 8.2% and 7.5% on VIPeR, iLIDS, PRID and Market, respectively; on CUHK01 we achieve competitive performance to CAMEL [46] (2.4% performance gap), but CAMEL performs significantly worse than our approach on VIPeR and Market. Compared with other augmentation techniques (*e.g.* SPGAN [49]), our illumination inference step ensures that the source illumination, chosen from a large number of options in our SyRI dataset, is closest to the target domain. (2) When compared to unsupervised hand-crafted based approaches, the performance margins for rank-1 are even larger: 11.5%, 13.9%, 7.2% and 18% on VIPeR, CUHK01, iLIDS and PRID, respectively. (3) Our approach is also very competitive with the best supervised techniques—regardless of the dataset. This confirms the effectiveness of the proposed solution, which does not require any human supervision and thus scales to large camera networks.

|   |   | METHOD | VIPeR | CUHK01 | iLIDS | PRID | Market |
|---|---|---|---|---|---|---|---|
| unsupervised | hand-craft | GL [18] | 33.5 | 41.0 | - | 25.0 | - |
|  |  | DLLAP [20] | 29.6 | 28.4 | - | 21.4 | - |
|  |  | TSR [19] | 27.7 | 23.3 | - | - | - |
|  |  | TL [48] | 31.5 | 27.1 | 49.3 | 24.2 | - |
|  | CNN | SSDAL [50] | 37.9 | - | - | 20.1 | 39.4 |
|  |  | CAMEL [46] | 30.9 | **57.3** | - | - | 54.5 |
|  |  | SPGAN [49] | - | - | - | - | 57.7 |
|  |  | TJ-AIDL [47] | 38.5 | - | - | 34.8 | 58.2 |
|  |  | **Ours** | **43.0** | 54.9 | **56.5** | **43.0** | **65.7** |
| supervised | hand-craft | LOMO+XQDA [4] | 40.0 | 63.2 | - | 26.7 | - |
|  |  | Ensembles [45] | 45.9 | 53.4 | 50.3 | 17.9 | - |
|  |  | Null Space [51] | 42.2 | 64.9 | - | 29.8 | 55.4 |
|  |  | Gaussian+XQDA [5] | 49.7 | 57.8 | - | - | 66.5 |
|  | CNN | Triplet Loss [52] | 47.8 | 53.7 | 60.4 | 22.0 | - |
|  |  | FT-JSTL+DGD [8] | 38.6 | 66.6 | 64.6 | 64.0 | 73.2 |
|  |  | SpindleNeT [6] | <span style="color:red">53.8</span> | <span style="color:red">79.9</span> | <span style="color:red">66.3</span> | <span style="color:red">67.0</span> | <span style="color:red">76.9</span> |

Table 3: **Performance comparison with state-of-the-art unsupervised and supervised techniques**. CMC rank-1 accuracies are reported. The best scores for unsupervised methods are shown in **bold**. The best scores of supervised methods are highlighted in <span style="color:red">red</span>. Our results are comparable with supervised methods that require hundreds or thousands of labeled images across camera pairs for training.

## 6  Conclusion

Re-identification datasets contain many identities, but rarely have a substantial number of different lighting conditions. In practice, this lack of diversity limits the generalization performance of learned re-identification models on new unseen data. Typically, the networks must be fine-tuned in a supervised manner using data collected for each target camera pair, which is infeasible at scale. To solve this issue, we propose a new synthetic dataset of virtual people rendered in indoor and outdoor environments. Given example unlabelled images from a test camera, we develop an illumination condition estimator to select the most appropriate subset of our synthesized images to use for fine-tuning a pre-trained re-identification model. Our approach is ideal for large scale deployments, since no labelled data needs to be collected for each target domain.

We employ a deep network to modify the subset of synthesized images (selected by the illumination classifier) so that they more closely resemble images from the test domain (see Fig. 6). To accomplish this, we use the recently introduced cycle-consistent adversarial architecture and integrate an additional regularization term to ensure the learned domain shift (between synthetic and real images) does not result in generating unrealistic training examples (*e.g.* drastic changes in color). Because re-identification images have distinct foreground/background regions, we also incorporate a soft matte to help the network focus on ensuring the foreground region is correctly translated to the target domain. Extensive experiments on multiple datasets (see Tab. 3) show that our approach outperforms other unsupervised techniques, often by a large margin.